\pdfoutput=1
\documentclass[11pt]{article}

\usepackage[]{EMNLP2023}

\usepackage{times}
\usepackage{latexsym}

\usepackage[T1]{fontenc}
\usepackage[utf8]{inputenc}
\usepackage[utf8]{inputenc}
\usepackage[noend]{algpseudocode}
\usepackage{varwidth}
\usepackage{tabularx}
\usepackage{subcaption}
\usepackage{amssymb}
\usepackage{amsmath}
\usepackage{amsthm}

\usepackage{mathabx}
\usepackage{amsfonts}
\usepackage{bbm}
\usepackage{bm}
\usepackage{booktabs}
\usepackage{multirow}
\usepackage{multicol}
\usepackage{mdwlist}
\usepackage{csquotes}
\usepackage{xcolor}
\usepackage{graphicx}
\usepackage{graphicx}
\usepackage{booktabs}
\usepackage{graphics}
\usepackage{tabularx}
\usepackage{xspace}
\usepackage{booktabs}
\usepackage{mdwlist}
\usepackage{mathtools}
\usepackage{enumitem}
\usepackage{calc}
\definecolor{Gray}{gray}{0.85}
\usepackage[neverdecrease]{paralist}
\usepackage[ruled,noend]{algorithm2e}
\usepackage{booktabs}
\usepackage{diagbox}
\usepackage{colortbl}
\usepackage{subcaption}
\usepackage{makecell}
\usepackage{verbatim}
\usepackage{amsmath}

\usepackage{color,soul}

\usepackage{microtype}

\usepackage{inconsolata}

\title{How much reliable is ChatGPT's prediction on Information Extraction under Input Perturbations?}

\author{Ishani Mondal\thanks{Equal Contribution}\\
  University of Maryland, College Park \\
  College Park, Maryland, USA \\
  \texttt{imondal@umd.edu} \\\And
  Abhilasha Sancheti$^{*}$ \\
  University of Maryland, College Park \\
  College Park, Maryland, USA \\
  \texttt{sancheti@umd.edu} \\}

\begin{document}
\maketitle
\begin{abstract}
In this paper, we assess the robustness (reliability) of ChatGPT under input perturbations for one of the most fundamental tasks of Information Extraction (IE) i.e. Named Entity Recognition (NER). Despite the hype, the majority of the researchers have vouched for its language understanding and generation capabilities; a little attention has been paid to understand its robustness:  How the input-perturbations affect 1) the predictions, 2) the confidence of predictions and 3) the quality of rationale behind its prediction. We perform a systematic analysis of ChatGPT's robustness (under both zero-shot and few-shot setup) on two NER datasets using both automatic and human evaluation. Based on automatic evaluation metrics, we find that 1) ChatGPT is more brittle on \textbf{Drug} or \textbf{Disease} replacements (rare entities) compared to the perturbations on widely known \textbf{Person} or \textbf{Location} entities, 2) the quality of explanations for the same entity \textit{considerably differ} under different types of \enquote{Entity-Specific} and \enquote{Context-Specific} perturbations and the quality can be significantly improved using in-context learning, and 3) it is overconfident for majority of the incorrect predictions, and hence it could lead to misguidance of the end-users\footnote{A short version accepted ICLR Tiny Paper 2024 is available here \url{https://openreview.net/pdf?id=cyN5Ck1RFT}}.

\end{abstract}

\section{Introduction}
\label{sec:intro}

\begin{figure}[!t] 
    \centering
    \includegraphics[width=\linewidth]{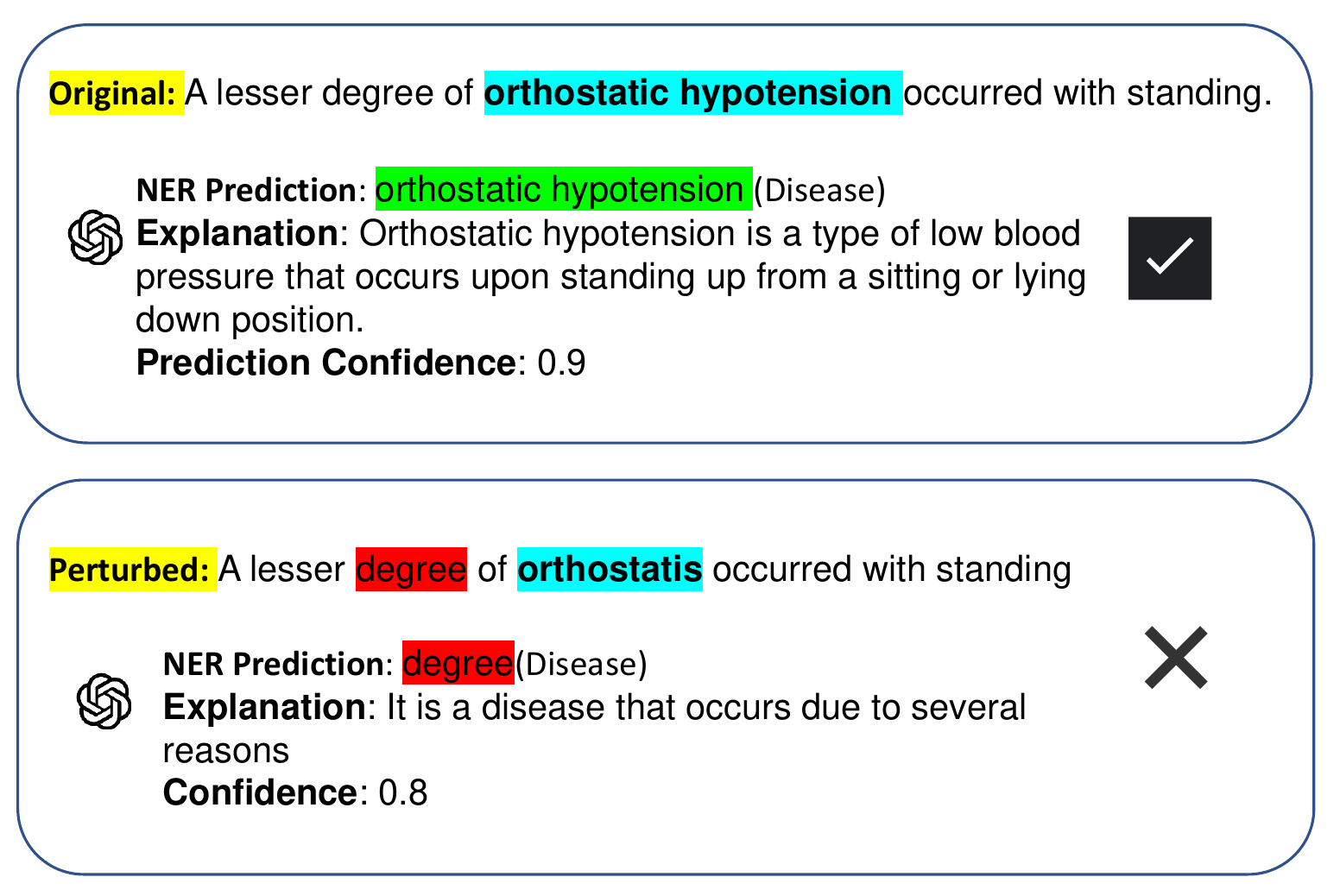}
    \caption{An example of sentence from BC5CDR in which the disease entity \textbf{orthostatic hypotension} has been perturbed with a synonym  \textbf{orthostatis}. Before perturbation, the disease was correctly predicted and explained with high confidence (90\%). After perturbation, \textit{degree} has been incorrectly predicted as a disease entity with a wrong explanation. However, ChatGPT is nearly equally confident (80\%) as the situation when it made a correct prediction.}
    \label{fig:diagram}
\end{figure}

The rapidly evolving field of natural language processing (NLP) witnesses the upsurge of large language models (LLMs) (such as (\textit{e.g.}, GPT3~\citep{brown2020language}, LaMDA~\citep{thoppilan2022lamda} and PaLM~\citep{chowdhery2022palm}, etc.).
Prompting these models has emerged as a widely adopted paradigm, given their superior zero-shot learning capability \citep{min2022rethinking}. Moreover, with proper instruction \cite{hegselmann2023tabllm, ma2023fairnessguided}, these LLMs achieve better performances on downstream NLP tasks. ChatGPT\footnote{https://openai.com/blog/chatgpt} is one such powerful LLM that has attracted a huge volume of users ever since its inception.  However, it is unclear whether it is \textit{reliable} in the realistic applications in which entities or context words can be out of the distribution of the training data, thereby calling attention to gauge its robustness. While previous efforts have evaluated various aspects of ChatGPT in law~\citep{choi2023chatgpt}, ethics~\citep{shen2023chatgpt}, education~\citep{khalil2023will}, verifiability~\citep{liu2023evaluating} and reasoning~\citep{bang2023multitask}, we focus on its robustness~\citep{bengio2021deep} to input perturbations, which has not been thoroughly evaluated yet. 

Since ChatGPT is a black-box model that hardly provides any information about its training details, the generated responses can significantly influence user’s trust \citep{deshpande2023toxicity, 10.1145/3543873.3587368}. Hence the evaluation based on its sensitivity to input perturbations should also involve gauging the \textit{reliability} of responses \textbf{under the light of robustness} by investigating its \textbf{prediction confidence} and the \textbf{rationale}\footnote{We use the terms \textit{rationale} and \textit{explanation}, \textit{robustness} and  \textit{reliability} interchangably} behind its prediction.

To assess that, we focus on an elementary IE task, \textit{i.e.}, Named Entity Recognition (NER). We make changes in the input data at both \emph{``Entity-level''} (by replacing target entities with other entities of the \textbf{same semantic class in Wikidata}, \textbf{typo}, \textbf{alias}, \textbf{random string}''); and at the \emph{``Context-level''} (by using pre-trained language models (e.g., BERT \cite{devlin-etal-2019-bert}) to generate contextual \textbf{verb substitutions}).
In this paper, we investigate the \textit{reliablity} of ChatGPT's prediction under both zero-shot and few-shot settings along three dimensions, including 1)
\textbf{Performance Shift with Perurbation:} How does the robustness of ChatGPT vary with domains and types of perturbations? Does ChatGPT make incorrect predictions for the examples that are easy for humans?
2) \textbf{Difference in Explanation Quality with Perturbation:} Is ChatGPT better at explaining its predictions on a \emph{local-level} (grounded in the input) or a \emph{global-level} (grounded in world knowledge) and how does that vary with perturbations under zero-shot setup? We also examine if the perturbation in the target entity causes a change in the semantic similarity of other non-target entities' explanations before and after the perturbation. We have also assessed the difference in explanation quality of the same entities (\emph{local-level} or \emph{global-level}) before and after the perturbation using both automatic and human evaluation.
3) \textbf{Variation in Confidence Calibration with Perturbation}: Is there any difference between prediction confidence between correct and incorrect predictions due to perturbation?
%Evaluation Criteria, Robustness and Error Types.
 %and explanations by analyzing the effect of perturbation on the target entity and also other non-target entities.
%Our systematic analysis is pivoted on three primary research questions: \textbf{RQ1)} How does the robustness of ChatGPT vary with domains and types of perturbations? \textbf{RQ2)} Is ChatGPT better at explaining its predictions on a \emph{local-level} (grounded in the input) or a \emph{global-level} (grounded in world knowledge) and how does that vary under attack? \textbf{RQ3)} Does ChatGPT make incorrect predictions for the examples which are easy for humans? %\textbf{RQ4)} What is the correlation between 

% With an aim to answer these research questions, we perform automatic evaluation 

% : 1) ChatGPT’s overall performance of NER under adversarial attack, 2) if perturbed input causes a change in the way we determine if the explanations are grounded to the input, or if they are spurious, 3) if the perturbation in the target entity causes a change in the semantic similarity of other non-target entities explanation before and after the attack. We have also assessed the difference in explanation quality of the same entities (\emph{local-level} or \emph{global-level}) before and after attack using both automatic and human evaluation.

In a nutshell, our contributions are four fold:\\
\textbf{1.} To the best of our knowledge, we are the first to comprehensively analyze the effect of input perturbations on ChatGPT's predictions and the rationale behind its prediction.\\
\textbf{2.} Our automatic evaluation reveals that under the light of robustness, ChatGPT's predictions and faithfulness of explanations are less reliable on domain-specific entities compared to popular entities; and the quality of explanations for the overlapping entities which are predicted both before and after the perturbation also considerably vary, indicating less \textit{reliability}. \\
\textbf{3.} Human evaluation further validates our findings from automatic evaluation and we throw some light on human's notion of informativeness of explanations, ease of entity prediction under perturbations and how does that correlate with the behavior of ChatGPT.\\
\textbf{4.} Even though ChatGPT is overconfident for incorrect predictions, its overconfidence can be significantly reduced using in-context learning; the quality of explanations (containing both local and global cues) also improves under a few-shot setup.

\begin{table*}[!t]
\centering
    \tiny
    \resizebox{\textwidth}{!}{
    \scalebox{0.9}{\begin{tabular}{p{15mm} p{65mm} p{65mm}}
    \Xhline{1.5pt}
    \rowcolor{Gray}
     \textbf{Perturbations} & \textbf{Original Sentence ($S$)} & \textbf{Perturbed Sentence ($S'$)} \\[1.1ex] \Xhline{1.5pt}
    Same Entity Type & We tested the sulfated polysaccharide fucoidan , which has been reported to reduce inflammatory \colorbox{green!15}{brain damage}, in a rat model of intracerebral hemorrhage induced by injection of bacterial collagenase into the caudate nucleus . & We tested the sulfated polysaccharide fucoidan , which has been reported to reduce inflammatory \colorbox{red!15}{chorioretinal atrophy}, in a rat model of intracerebral hemorrhage induced by injection of bacterial collagenase into the caudate nucleus .\\
    \midrule
    Alias & CONCLUSION : This study confirms our previous finding that \colorbox{green!15}{selegiline}  in combination with L - dopa is associated with selective orthostatic hypotension . & CONCLUSION : This study confirms our previous finding that \colorbox{red!15}{l-deprenalin} in combination with L - dopa is associated with selective orthostatic hypotension . \\
    \midrule
    Typo & China on Thursday accused Taipei of spoiling the atmosphere for a resumption of talks across the Taiwan Strait with a visit to Ukraine by Taiwanese Vice President \colorbox{green!15}{Lien Chan} this week that infuriated Beijing .& China on Thursday accused Taipei of spoiling the atmosphere for a resumption of talks across the Taiwan Strait with a visit to Ukraine by Taiwanese Vice President \colorbox{red!15}{en ChanLi} this week that infuriated Beijing .\\
    \midrule
    Random & \colorbox{green!15}{Rabinovich} is winding up his term as ambassador & \colorbox{red!15}{I3qk2ia} is winding up his term as ambassador\\
    \midrule
    Verb & 
    Speaking only hours after Chinese state media \colorbox{green!15}{said} the time was right to engage in political talks with Taiwan , Foreign Ministry spokesman Shen Guofang told Reuters : " The necessary atmosphere for the opening of the talks has been disrupted by the Taiwan authorities . "& Speaking only hours after Chinese state media \colorbox{red!15}{announced} the time was right to engage in political talks with Taiwan , Foreign Ministry spokesman Shen Guofang told Reuters : " The necessary atmosphere for the opening of the talks has been disrupted by the Taiwan authorities . "
     
\\ \Xhline{1.5pt}
    \end{tabular}}}
    \caption{Examples of original sentences containing \colorbox{green!15}{target entities} ($T_E$) and the corresponding sentences with \colorbox{red!15}{perturbed entities}($T'_E$)  for both \enquote{\textit{Entity-Specific}} and \enquote{\textit{Context-Specific}} cases. These sentences are interpolated from CONLL and BC5CDR train datasets.}
    \label{tab:perturbation examples}
\end{table*}

\section{Can we automatically generate input perturbations?}
\label{sec:perturb}
Inspired by \cite{lin-etal-2021-rockner}, we generate high-quality examples for evaluating the robustness of ChatGPT on the task of NER by perturbing both the entities (\emph{\enquote{Entity-specific}}) and contexts (\emph{\enquote{Context-specific}}) of original examples. 
We refer to the perturbed entity as \emph{``target entity"} ($T_E$).
In a sentence ($S$) of length $n$, we denote a target entity as $T$ and it is replaced by a perturbating entity $T'_E$, thereby generating a perturbed sentence ($S'$). Besides, target entity there could be other possible $k$ entities ($O_E$ = {$O_{E_1}, O_{E2}..........O_{E_k}$}) (where $k$ $<$ $n$). Some samples of perturbed sentences are presented in Table~\ref{tab:perturbation examples}. It is important to note that, we perform perturbation of 1 target entity or verb at a time to generate $S'$ before checking NER prediction by ChatGPT.\\
\textbf{A. Entity-Specific:} In this case, we are generating the following perturbations of entities present in the sentences (containing $T_E$), and asking ChatGPT to predict named entities for the perturbed sentences (containing $T'_E$).\\
\textbf{a) Alias Replacement:} We use Wikidata API to link the target entity $T_E$ in original examples from its surface to canonical form in Wikidata with a unique identifier (\textbf{Entity Typing}) and generate $p$ aliases ($T_{Ea_{1}}$, $T_{Ea_{2}}$... $T_{Ea_{p}}$) of those entities. \\
\textbf{b) Same Entity Type Replacement:} We perturb $T_E$ with another entity of similar semantic class (For instance, a disease replaced by another disease). For this, we retrieve $p$ additional entities occurring in other input sentences. Then we perform $p$ replacements.\\
\textbf{c) Typo Replacement:} We also consider perturbing the target entity $T_E$ with natural-looking typos, such as the rotation of characters in the token of $T_E$.\\
\textbf{d) Random Entity Replacement:} We also replace target entity $T_E$ with one randomly generated string and hypothesize that the model would be able to detect the entity based on contextual cues.

\paragraph{B. Context-Specific:} Here we generate perturbations of the context around target entities and ask ChatGPT to predict named entities for the perturbed sentences which contain $T_E$, and perturbed contextual cues.\\
\textbf{Verb substitution with synonyms:} We generate \textit{context-specific} perturbations by substituting the main verb $v$ in the sentence with three synonyms ($v'_s1$, $v'_s2$, $v'_s3$) predicted by a pre-trained masked language model like BERT \cite{devlin-etal-2019-bert}.

\section{Experimental Setup}

\paragraph{Datasets:} 
We evaluate the explainability and NER capability of ChatGPT on CONLL-2003~\citep{tjong-kim-sang-de-meulder-2003-introduction} and BC5CDR~\citep{li2016biocreative} datasets by prompting ChatGPT (see \ref{sec:prompt} for prompt) to obtain the predicted entities and corresponding explanations in a structured format. We only consider two types of entity predictions (PERSON, LOCATION) from the CONLL-2003 dataset.

\paragraph{Evaluation Criteria:}
We provide a comprehensive understanding of how we approach our research questions mentioned in Section~\ref{sec:intro} and evaluate the robustness of ChatGPT under input perturbations. On a high-level, we define the following evaluation criteria to measure the same:

\paragraph{1. Performance Difference under Perturbation:} Motivated by \cite{advglue2021, mondal-2021-bbaeg}, we comprehensively evaluate the overall performance of ChatGPT on the NER task and compare it when the inputs are perturbed. By examining the change in its performance across two situations, we seek to provide a detailed understanding of ChatGPT’s reliability under input perturbations.

\paragraph{2. Difference in Quality of Explanations due to perturbation:} The explainability of ChatGPT is crucial for its application in real-world scenarios \cite{aghajanyan-etal-2021-intrinsic, rajani-etal-2019-explain}; hence we ask ChatGPT to provide reasons for its predictions before and after the perturbation. 

\paragraph{3. Confidence Calibration under Perturbation:}
Calibration measurement is a crucial aspect of determining the predictive uncertainty of a model \cite{guo2023close, ulmer2022exploring}; a well-calibrated classifier must have predictive scores that reflect the probability of its correctness \cite{NEURIPS2021_8420d359, NEURIPS2019_36ad8b5f}. Here we aim to identify the uncertainties and confidence of ChatGPT (by prompting it to provide a confidence score in the range of 0 to 1) in predicting named entities before and after perturbations.

\paragraph{Zero-shot and Few-shot Setup:}
To conduct a thorough evaluation of ChatGPT’s capabilities on the NER task, we first measure its performance in the \textbf{Zero-shot scenario}. 
Then, we investigate how a \textbf{Few-shot Approach} or in-context learning (ICL) approach affects its performance. 
First, we manually design different zero-shot prompts since ChatGPT is sensitive to different prompts, and then we choose the ones that provide maximally correct output on the non-perturbed sentences ($S$). 
Then we construct few-shot ICL prompts by selecting zero-shot prompts and randomly adding some samples from the corresponding training set. 
%While adding the samples, we also
%For few-shot COT prompts, we add the chain-of-thought explanations to the few-shot ICL prompts, where the chain-ofthought explanations are manually constructed with the help of ChatGPT. To eliminate the randomness of selected samples, we select five different groups and also report the means and standard deviations.

\paragraph{How are the prompts designed?}
The prompts designed for zero-shot settings consist of the following integral main elements: the task instruction, candidate target labels, output format description, and the input text. The task instruction describes the specific IE sub-task where we ask the model to provide confidence of its prediction and the explanation behind its prediction; candidate target labels are the types of target information, such as entity types; the output format description specifies the format of outputs to facilitate easy parsing. In the few-shot setting, we also provide some demonstration examples, which can also provide the chain-of-thought explanation and confidence of prediction. Since we assume a combination of local+global explanations are the most useful ones for NER prediction, we combine both wikipedia descriptions of the entities and local contextual cues in the explanation behind predicting it as an entity. (See Appendix~\ref{sec:prompt} for prompts used)

%Explain Zero-shot and Few-shot scenario
\subsection{Implementation Details}
We use \enquote{\emph{gpt-3.5-turbo}} model using OpenAI API key to obtain predictions for named entities and corresponding explanations for examples from the train-split for which triggers were collected by~\citet{lin2020triggerner}. For each of the examples, we generate $3$ perturbations per ground truth entity for \textbf{Alias, Verb, and Same Entity Type}, and $1$ for \textbf{Random Entity, and Typo}\footnote{Only 1 perturbation since it cannot have much variations}. 
To eliminate the randomness of predicted samples, we set the temperature to 0.
% \section{Experimental Setup}
%  We evaluate the explainability and NER capability of ChatGPT on CONLL-2003~\citep{tjong-kim-sang-de-meulder-2003-introduction} and BC5CDR~\citep{li2016biocreative} datasets by prompting ChatGPT (see \ref{sec:prompt} for prompt) to obtain the predicted entities and corresponding explanations in a structured format. 

% \subsection{Implementation Details}
% We use \enquote{\emph{gpt-3.5-turbo}} model using OpenAI API key to obtain predictions for named entities and corresponding explanations for examples from the train-split for which triggers were collected by~\citet{lin2020triggerner}. For each of the examples, we generate $3$ perturbations per ground truth entity for \textbf{Alias, Verb, and Same Entity Type}, and $1$ for \textbf{Random Entity, and Typo}\footnote{Only 1 perturbation since it cannot have much variations}.

% \subsection{Datasets}

% \label{tab:robustness}
% \end{table*}
\begin{table*}[t]
\small
\centering
\begin{tabular}{ l c c c c  c c} 
\toprule
&  \multicolumn{3}{c}{Effect on Target Entity}  & \multicolumn{2}{c}{Effect on non-target Entities} & Overall Effect \\
\cmidrule(r){2-4}
\cmidrule(r){5-6}
& $\Delta$ Accuracy & $\Delta$ Faithfulness & Similarity & $\Delta$ F1 & $\Delta$ Faithfulness & $\Delta$ F1\\
\midrule
\textit{Entity-Level} \\
Alias Perturbation & 0.16 / 0.03 & 0.10 / 0.05 & 0.69 / 0.81 & -0.13 / 0.01 &  0.01 / 0.01 & 0.01 /0.01 \\
Entity Type Perturbation & 0.10 / 0.15 & 0.09 / 0.08 & 0.58 / 0.74 & 0.03 / 0.02 & 0.03 / 0.03 & 0.03 / 0.02 \\
Typo Perturbation & 0.30 / 0.13 & 0.21 / 0.15 & 0.63 / 0.76 & 0.01 / 0.01 &  0.01 / 0.01 & 0.04 / 0.03\\
Random Perturbation & 0.38 / 0.20  & 0.27 / 0.15 & 0.49 / 0.79  & 0.02 / 0.01 &  0.01 / 0.01 &  0.08 / 0.06 \\
\midrule
\textit{Context-Level} \\
Verb Substitution & - & - & - &  0.01 / 0.01 & 0.01 / 0.01 & 0.02 / 0.02 \\
 \bottomrule

\end{tabular}
\caption{
Assessment of Robustness of NER predictions, Faithfulness of its predictions to input (extrinsic), and similarity between the explanation generated for the original and perturbed instances in the form of (zero-shot / few-shot) prediction performances on the BC5CDR Dataset. 
}
\label{tab:robustness-bc5cdr}
\end{table*}

\begin{table*}[t]
\small
% \setlength\tabcolsep{5pt}
%\small
\centering
\begin{tabular}{ l c c c c  c c} 
\toprule
&  \multicolumn{3}{c}{Effect on Target Entity}  & \multicolumn{2}{c}{Effect on non-target Entities} & Overall Effect \\
\cmidrule(r){2-4}
\cmidrule(r){5-6}
& $\Delta$ Accuracy & $\Delta$ Faithfulness & Similarity & $\Delta$ F1 & $\Delta$ Faithfulness & $\Delta$ F1\\
\midrule
\textit{Entity-Level} \\
Alias Perturbation & 0.06 / 0.03 & 0.03 / 0.02 & 0.77 / 0.78  & 0.01 / 0.01  &  0.03  / 0.03 & 0.03 / 0.03 \\
Entity Type Perturbation & 0.06 / 0.04 & 0.06 / 0.05 & 0.75 / 0.82  & 0.01 / 0.005 & 0.02 / 0.01  &  0.02 / 0.01  \\
Typo Perturbation & 0.54 / 0.33 & 0.46 / 0.24  & 0.37 / 0.46 & 0.03 / 0.02 &  0.01 / 0.01 & 0.05 / 0.04 \\
Random Perturbation & 0.23 / 0.11 & 0.15 / 0.09 & 0.60 / 0.64 &  0.02 / 0.02 & 0.02 / 0.02 &  0.07 / 0.07 \\
\midrule
\textit{Context-Level} \\
Verb Substitution & - & - & - &  0.01 / 0.01 & 0.02 / 0.01 & 0.01 / 0.02 \\
 \bottomrule
\end{tabular}
\caption{
Assessment of Robustness of NER predictions, Faithfulness of its predictions to input (extrinsic), and similarity between the explanation generated for the original and perturbed instances in the form of (zero-shot / few-shot) prediction performances on the CONLL dataset. 
}
\label{tab:robustness-conll}
\end{table*}

%For each type of perturbation, we \textbf{highlight} the maximum value between CONLL or BC5CDR metrics to reflect on the effect of perturbation on domain.

\begin{table*}[t]
\small
\centering
\begin{tabular}{ l c c c c | c c c c} 
\toprule
&  \multicolumn{4}{c|}{Global vs Local Explanations (Zero-shot)} & \multicolumn{4}{c}{Global vs Local Explanations (Few-shot)}\\
\cmidrule(r){2-5} \cmidrule(r){6-9}
& G$\upuparrows$L$\upuparrows$ & G$\downdownarrows$L$\upuparrows$ & G$\upuparrows$L$\downdownarrows$ & G$\downdownarrows$L$\downdownarrows$ & G$\upuparrows$L$\upuparrows$ & G$\downdownarrows$L$\upuparrows$ & G$\upuparrows$L$\downdownarrows$ & G$\downdownarrows$L$\downdownarrows$ \\
\midrule
\textit{BC5CDR} \\
Alias  & \textbf{0.54} & 0.26 & 0.17 & 0.02  & \textbf{0.57} & 0.20 & 0.18 & 0.03  \\
Same Entity Type & \textbf{0.61} & 0.22 & 0.13 & 0.02 & \textbf{0.48} & 0.29 & 0.16 & 0.06 \\
Typo & \textbf{0.36} & 0.26 & 0.26 &  0.10 & \textbf{0.46} & 0.19 & 0.30 &  0.03\\
Random & 0.39 &  \textbf{0.43} & 0.17 &  0.00 & \textbf{0.46} &  0.15 & 0.19 &  0.19 \\
Verb & 0.24 & \textbf{0.48} & 0.24 & 0.02 & \textbf{0.48} & 0.28 &   0.20 & 0.02 \\
\midrule    
\textit{CONLL} \\
    Alias  & 0.21 & \textbf{0.58} & 0.06 & 0.13  & \textbf{0.60} & 0.24 & 0.05 & 0.11\\
Same Entity Type & 0.21 & \textbf{0.48} & 0.15 & 0.16 & \textbf{0.46} & 0.23 & 0.15 & 0.16\\
Typo & 0.24 & \textbf{0.40} & 0.15 &  0.20  & \textbf{0.34} & 0.30 & 0.15 &  0.20 \\
Random & 0.11 & \textbf{0.63} & 0.15 &  0.10  & \textbf{0.60} & 0.20 & 0.20 &  0.11 \\
Verb & 0.22 & \textbf{0.56} & 0.07 & 0.13 & \textbf{0.56} & 0.22 & 0.07 & 0.13\\
 \bottomrule
\end{tabular}
\caption{Shows the change in the generated explanations due to the predictions of common entities before and after the perturbation. Here $\upuparrows$ and $\downdownarrows$ indicate increase and decrease after perturbation respectively.}\label{tab:explainability-analysis}
\end{table*}

\section{How to estimate \textit{Reliability}?}
We perform both \textbf{automatic} and \textbf{manual} evaluations of the predictions and generated explanations separately for target ($t$) (in ~\ref{para:target}), non-target ($nt$) (in ~\ref{para:nontarget}), and overall entities ($T$)\footnote{We consider entities to be case-sensitive for accuracy computation as NER can be considered as a span (grounded in input) prediction task}. 
Based on \textbf{automatic} evaluation, we come up with the following evaluation metrics that align well with answering our research questions (as laid out in \textsection\ref{sec:intro}): 

\subsection{Automatic Evaluation}
\paragraph{Is there any effect of ChatGPT's NER prediction on the target entity?}
We hypothesize that after each type of perturbation (\textsection\ref{sec:perturb}), we can observe some differences in the reliability of predicting target entities along with explanations provided in support of its predictions. In this case, we generate $S'$ \label{para:target} by perturbing $T_E$ with $T'_E$, and evaluate if ChatGPT can predict $T'_E$ correctly in $S'$, since from the contextual cues a smart human can predict the entity correctly instead of just predicting based on prior knowledge about the entity.

\paragraph{A: Accuracy Before and After Perturbation:} For each type of perturbation, we measure the difference in the accuracy of predicting the target entity $T_E$ before prediction and $T'_E$ after perturbation ($\Delta$ Accuracy). This is measured with respect to gold annotated entities in train split. Lesser the $\Delta$ Accuracy, higher is the robustness. 

\paragraph{B: Faithfulness of Explanation Before and After Perturbation:} Ideally, entity prediction based on contextual cues should be \textit{faithful} to the input context even after the perturbation. Thus, we measure the difference in the faithfulness (\emph{local-level} explanation measured in terms of cosine similarity of explanation with the input query) of explanation for the target entity prediction before ($T_E$) and after ($T_E$) perturbation ($\Delta$ Faithfulness). 

\paragraph{C: Similarity of Explanation Before and After Perturbation:} We measure the cosine similarity of the explanation generated for the prediction of the target entity before ($T_E$) and after ($T_E$) perturbations.

\paragraph{Is there any effect of ChatGPT's NER prediction on the non-target entities?}
\label{para:nontarget}
Here we aim to analyze whether after perturbing $T_E$ with $T'_E$, ChatGPT's predictions on $O_E$ alter. In other words, our primary goal is to verify if ChatGPT can successfully ignore the target entity perturbation and generate similar predictions and explanations for the other entities in $S$ and $S'$. Here also a smart human can predict the other non-perturbed entities based on contextual cues.

\paragraph{A: F1-Score Before and After Perturbation:} We measure the difference in the F1 of the prediction of non-target entities ($O_E$) (with respect to gold standard annotations) before and after perturbation ($\Delta$ F1). Lesser the $\Delta$ F1, higher is the robustness. 

\paragraph{B: Faithfulness of Explanation Before and After Perturbation:} 
% \paragraph{Extrinsic Evaluation:}
We measure the difference in the faithfulness (measured in terms of cosine similarity of explanation with the input query) of explanation for the target entity ($t$) prediction before and after perturbation ($\Delta$ Faithfulness).
For a certain type of perturbation in \textsection\ref{sec:perturb}, if there are $x$ inputs each containing $n$ entities on average, and each entity has $k$ different perturbations, then each of the above-mentioned metrics is reported using the weighted average rule: ($n$ * $k$)/$x$. We use these metrics to answer our \textbf{RQ1} in \textsection\ref{sec:intro}. 

We \textbf{approximate how the explanation of an entity is grounded to world knowledge (\emph{global-level})} by obtaining the entity description from wikipedia\footnote{\url{https://pypi.org/project/wikipedia/}} and calculating the similarity of generated explanation with respect to the summary.
Then \textbf{we analyze the effect of perturbations on generating \emph{global} and \emph{local} explanations} for the common non-target entities which are predicted both before and after perturbations in order to answer \textbf{RQ2}. Here we assume that whenever the explanation's faithfulness (both \emph{local} or \emph{global}) score changes for the same entity as before, we try to measure when there is an increase or decrease and enumerate those for both zero-shot and few-shot approach in Table~\ref{tab:explainability-analysis}.
% \paragraph{Intrinsic Evaluation:}
% We redefine the concept of \textit{faithfulness} with respect to input by performing an intrinsic evaluation. Since TriggerNER (containing both CONNLL and BC5CDR) dataset contains trigger (contextual cuess) annotation corresponding to each entity in the input sentence, we also measure the word overlap between the entity explanation provided by ChatGPT and the annotated triggers in order to quantify \textit{faithfulness} intrinsically.

\begin{table}[t]
% \tiny
% \setlength\tabcolsep{5pt}
\small
\centering
\begin{tabular}{ l c c c c} 
\toprule
&  \multicolumn{4}{c}{Confidence of Correct+Incorrect Predictions} \\
\midrule
&  \multicolumn{2}{c}{Zero-Shot} &  \multicolumn{2}{c}{Few-Shot}  \\
\cmidrule(r){2-5}
& $\Delta$C BP & $\Delta$C AP & $\Delta$C BP & $\Delta$C AP \\
\midrule
\textit{BC5CDR} \\
Alias & 0.12 & 0.05 & 0.18 & 0.09 \\
Typo & 0.10 & 0.08 & 0.15 & 0.13 \\
Random & 0.11 & 0.08 &  0.15 & 0.11 \\
Same Type & 0.07 & 0.11 &  0.14 & 0.18 \\
Verb & 0.04 & 0.08 &  0.01 & 0.03 \\
 \midrule
 \textit{CONLL} \\
Alias & 0.05 & 0.03 & 0.08 & 0.06 \\
Typo & 0.12 & 0.08 & 0.15 & 0.11 \\
Random & 0.21 & 0.18 &  0.23 & 0.20 \\
Same Type & 0.07 & 0.11 &  0.15 & 0.17 \\
Verb & 0.05 & 0.06 &  0.01 & 0.05 \\
\bottomrule
\end{tabular}
\caption{shows the change in the average confidence scores between the correct and incorrect predictions before (BP) and after perturbation (AP) in both zero-shot and few-shot predictions.}\label{tab:confidence}
\end{table}

\begin{figure*}[t]
    \centering
    \begin{subfigure}[t]{\columnwidth}
        \centering
    \includegraphics[width=0.8\columnwidth]{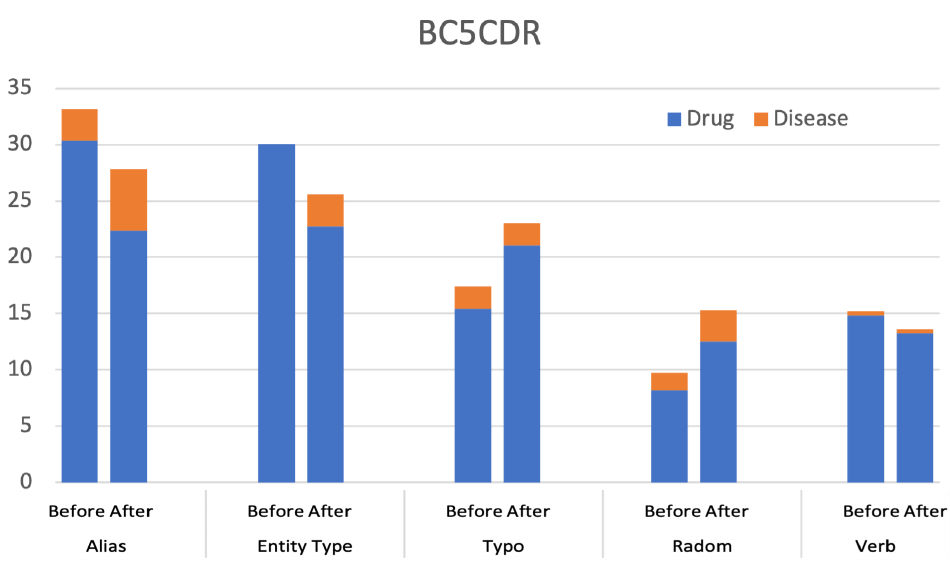}
        \caption{BC5CDR}\label{fig:bc5cdr}
    \end{subfigure}
    \begin{subfigure}[t]{0.8\columnwidth}
        \centering
        \includegraphics[width=\columnwidth]{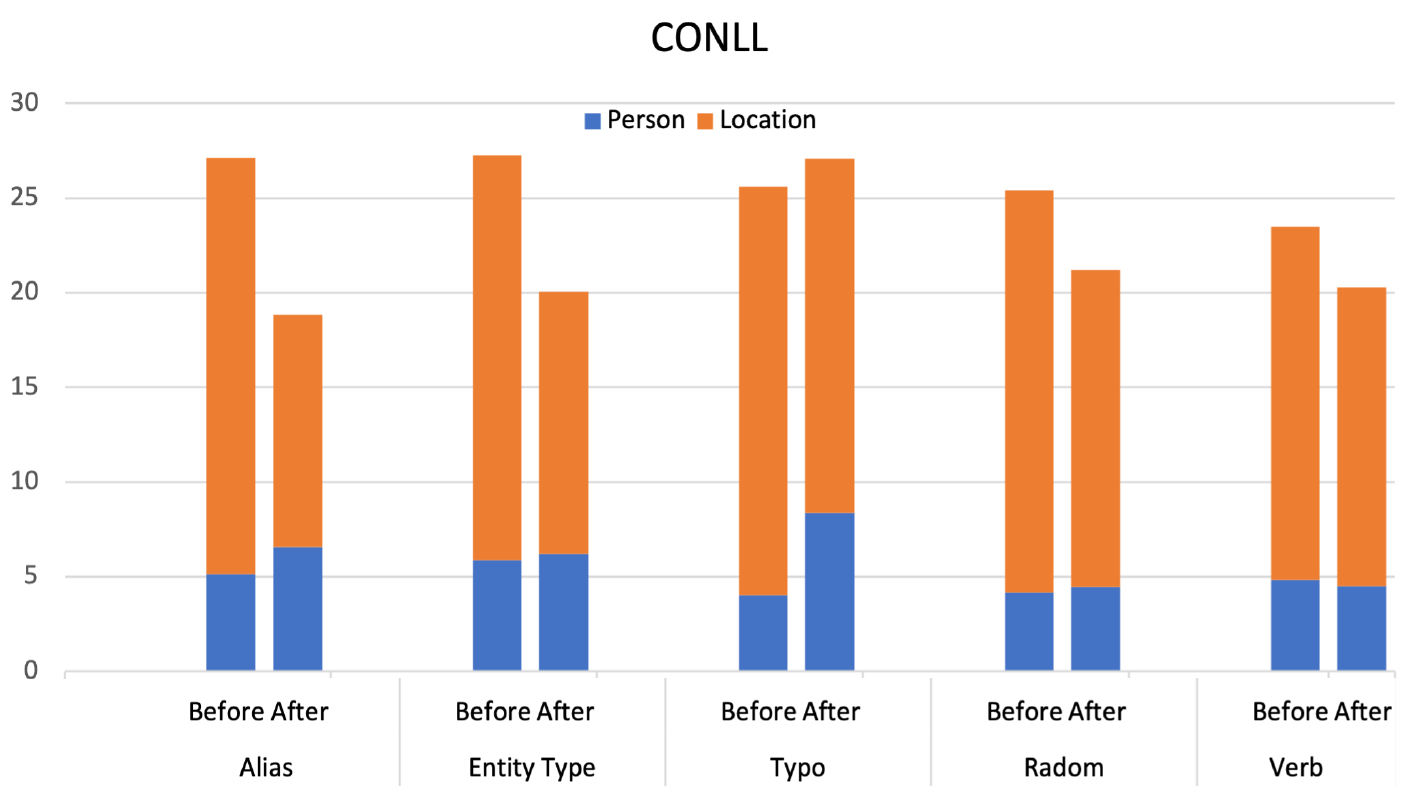}
    \caption{CONLL}\label{fig:conll}
    \end{subfigure}
    \caption{Percentage of examples \textbf{Before} and \textbf{After} perturbation for which the explanations are less informative such as ``refers to a country/person", ``it is a chemical compound/substance" for BC5CDR and CONLL datasets.} \label{fig:informativeness}
\end{figure*}
\paragraph{Confidence Calibration}
We estimate confidence in terms of a probability value (0-1) indicating the likelihood of belonging to a specific category, for both the correct and incorrect samples. We estimate overconfidence as the difference between confidence scores on correct and incorrect predictions ($\Delta$C). We aim to evaluate how $\Delta$C varies for different perturbations and if that gets reduced due to in-context learning.

After that, \textbf{Manual} analysis of the explanations is done for the target and non-target entities before and after perturbation. We sample $5$ inputs and their perturbations for each of the four possibilities (correct/incorrect prediction before/after the perturbation) to answer \textbf{RQ3} and further confirm findings obtained for \textbf{RQ2} from automatic evaluation.
% 
% \begin{table}[!t]
%     \renewcommand{\arraystretch}{1.2}
%     \centering
%     \small
%     \begin{tabular}{ccccc}
%     \Xhline{1pt}
%     && \multicolumn{2}{c}{\abr{dpr}}\\
%           \multirow{-2}{*}{} &&{Correct} & {Incorrect}
%          \\ 
%          \Xhline{1pt}
%          %\parbox[t]{2mm}{\multirow{2}{*}%{\rotatebox[origin=c]{90}{\textsc{Chat}}}} 
%          %\parbox[t]{2mm}{\multirow{2}{*}%{\rotatebox[origin=c]{90}{ \textsc{GPT}}}} 
%          \multirow{2}{*}{\abr{ChatGPT}}& Correct   &  2\%     & 33\% \\
%          & Incorrect &  1\%    & 64\% \\
%     \Xhline{1pt}
%     \end{tabular}
%     \caption{Percentage of questions stumping both \abr{dpr} and \abr{chatgpt} were the highest among all, indicating that the questions that were written with \abr{dpr} guidance also stumped \abr{chatgpt}.}
%     \label{tab:stumpaccuracy}
% \end{table}

\begin{figure*}[t]
    \centering
    \begin{subfigure}[t]{0.80\columnwidth}
        \centering    \includegraphics[width=\columnwidth]{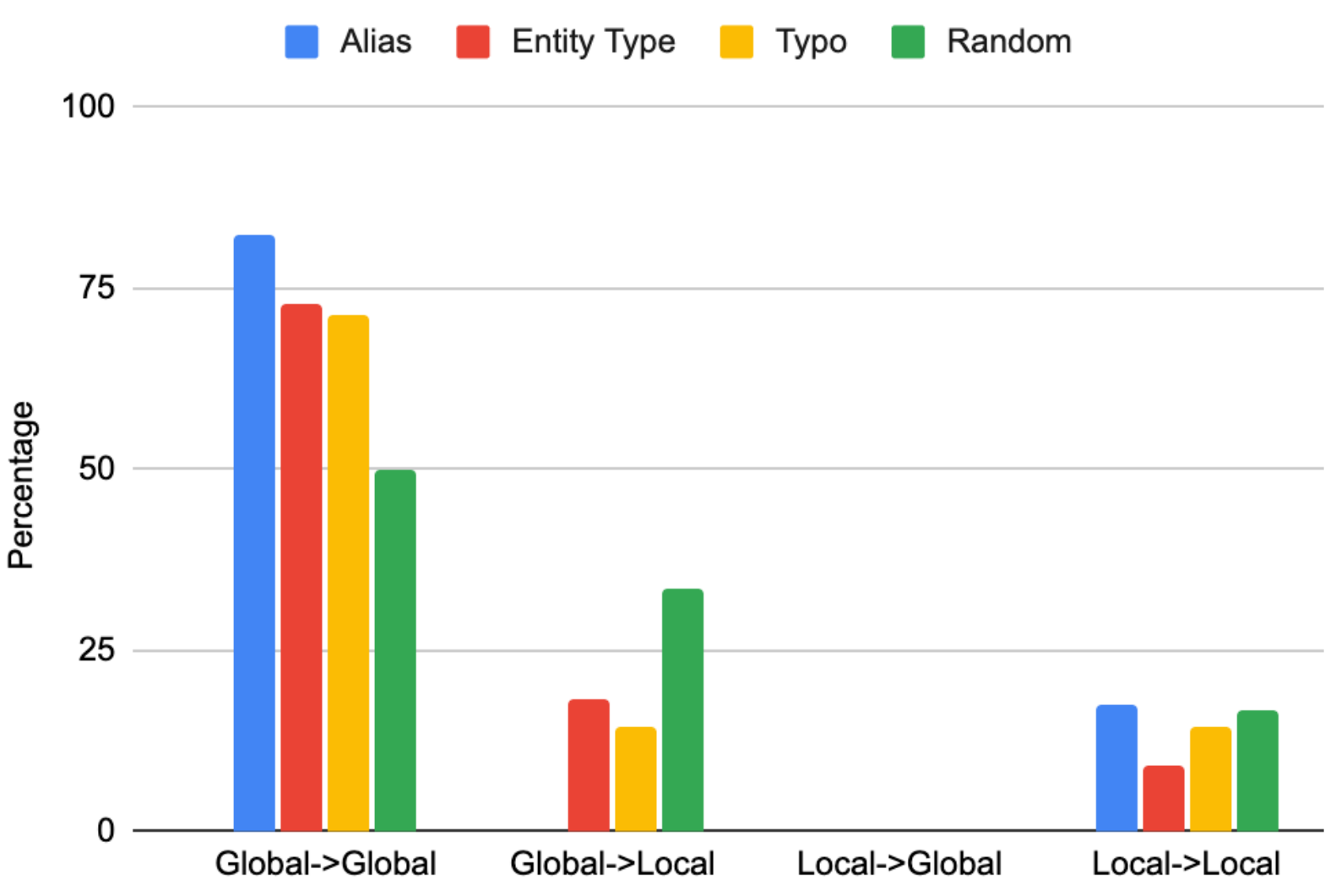}
        \caption{Target Entities}\label{fig:global-local-target}
    \end{subfigure}
    \begin{subfigure}[t]{0.80\columnwidth}
        \centering
        \includegraphics[width=\columnwidth]{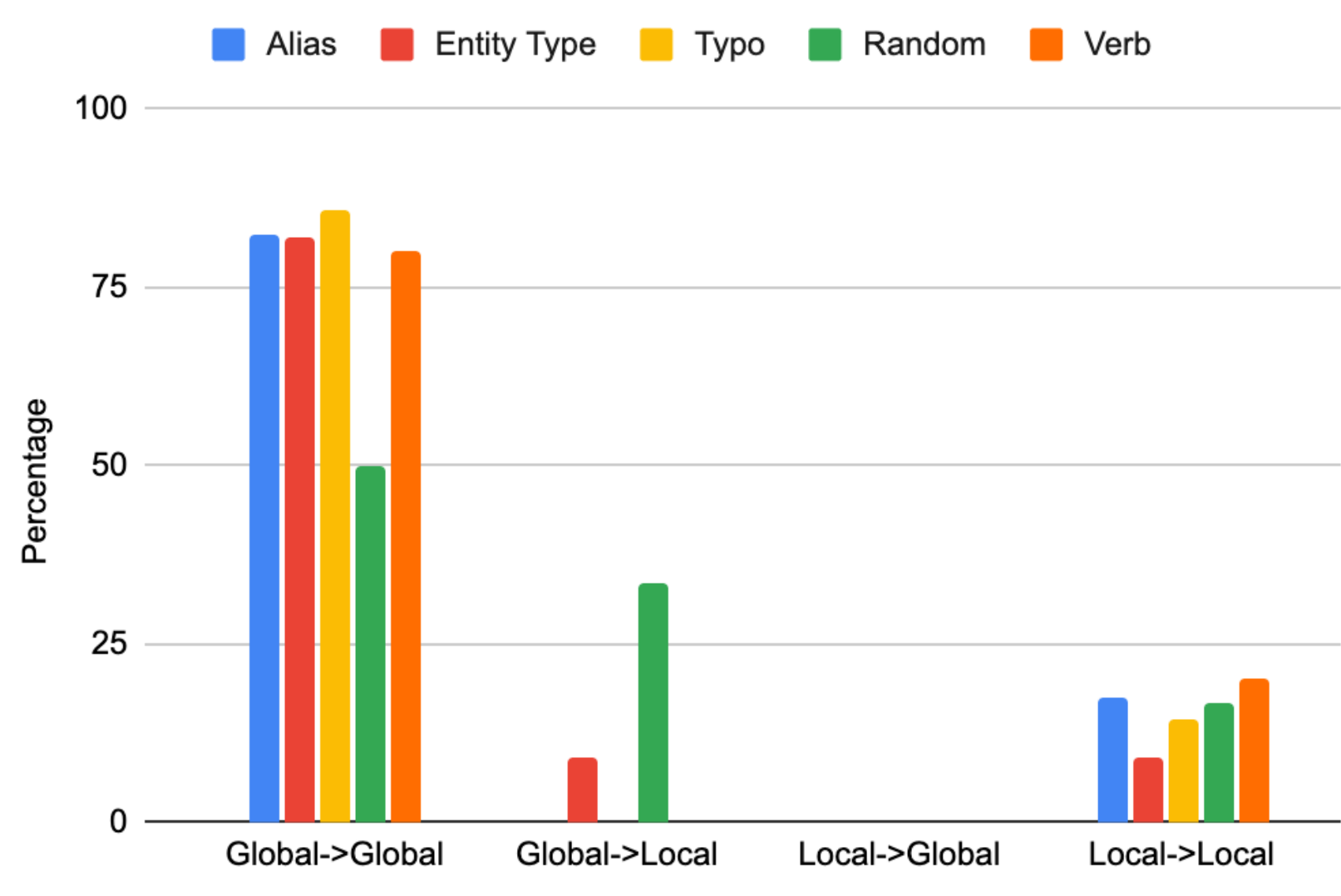}
    \caption{Non-Target Entities}\label{fig:global-local-other}
    \end{subfigure}
    \caption{Percentage of (input, perturbed input) pairs with change in type of explanations for (i) target and (ii) non-target entities in BC5CDR.} \label{fig:global-local-manual}
\end{figure*}

\section{Results and Findings}
\subsection{Automatic Evaluation}
\label{subsec:auto}
\paragraph{Robustness depends on perturbation type and domain of perturbing entities.}
Table~\ref{tab:robustness-bc5cdr} and Table~\ref{tab:robustness-conll} show that under  \textbf{zero-shot scenario} ChatGPT is more brittle on \textbf{Drug} or \textbf{Disease} replacements (rare entities) compared to the perturbations on widely known \textbf{Person} or \textbf{Location} entities in CONLL in terms of $\Delta$ Accuracy and $\Delta$ Faithfulness. Besides, Typo and Random entity substitution seems too brittle in terms of both these metrics. Using human evaluation, we wanted to confirm if the incorrectly predicted examples are also difficult to be identified by the humans. However, we notice that under   \textbf{few-shot scenario}, $\Delta$Accuracy gradually decreases for almost all the perturbations in both the datasets, indicating high robustness.

\paragraph{Transition of global and local explainability for same entity prediction under perturbation.} Based on the \textbf{zero-shot} results in Table~\ref{tab:explainability-analysis}, we observe that overall, the globality of explanations decreases while faithfulness to input increases due to perturbation. This provides us with an insight that when an entity is being perturbed, ChatGPT relies more on local context cues to detect entities. This holds true for all types of perturbations in CONLL since person or location names are widely popular, hence before perturbation major predictions were pivoted on world knowledge.
However, for Alias, Entity Type, Typo perturbations in BC5CDR, the explanations were more global and local before perturbation.
Thus, for the well-known entity types, the model chooses either local or global explanations, whereas after random perturbations, the models always prefer looking at contextual cues. Since while performing \textbf{few-shot} experiments, our goal has been to increase both locality and globality in all the explanations of the predicted entities (G$\upuparrows$L$\upuparrows$), we notice that the performance improves significantly under \textbf{few-shot as shown in Table~\ref{tab:explainability-analysis}}. Sample output predictions for sentences containing target entities in order to show the difference in the quality of explanations under zero-shot and few-shot setup are shown in Table~\ref{tab:few-shot-explanations}.

\paragraph{Overconfidence of ChatGPT on incorrect predictions can be alleviated to some extent using In-context Learning.} 
Table~\ref{tab:confidence} shows the difference in the average confidence scores of the correct and incorrect predictions before (BP) and after (AP) the perturbation in both zero-shot and few-shot predictions. It can be observed that under zero-shot scenario, ChatGPT is highly overconfident on the incorrect predictions, causing very little difference between correct and incorrect predictions. However, for all types of perturbations, the few-shot setup improves reliability in predictions even after the perturbation by increasing the gap between correct and incorrect predictions (except verb substitutions) ($\Delta$C). Moreover, $\Delta$C is being reduced after perturbations, indicating more overconfidence in incorrect predictions due to perturbation.

\begin{table*}[!t]
\centering
    \tiny
    \resizebox{\textwidth}{!}{
    \scalebox{0.9}{\begin{tabular}{p{50mm} p{55mm} p{55mm}}
    \Xhline{1.5pt}
    \rowcolor{Gray}
     \textbf{Original Sentence[Perturbation]} & \textbf{Predictions Before (Zero-shot)} & \textbf{Predictions Before (Few-shot)} \\[1.1ex] \Xhline{1.5pt}
     % & \textbf{Predictions Before} & \textbf{Predictions After} \\
    \colorbox{green!15}{Selegiline}[\colorbox{red!15}{L-deprenyl}]-induced postural hypotension in Parkinson's disease: a longitudinal study on the effects of drug withdrawal.&  \textbf{Chemical}: Selginline \textbf{Explanation}: Selegiline is a medication used to treat Parkinson's disease.
    \textbf{Confidence}: 0.8  & \textbf{Chemical}: Selginline \textbf{Explanation}: Selegiline is a chemical since it is a medication used to treat Parkinson's disease and it is mentioned in the sentence as a cause of postural hypotension.
    \textbf{Confidence}: 0.8  \\
    \midrule
    \colorbox{green!15}{Orthostatic hypotension}[\colorbox{red!15}{Orthostasis}] was ameliorated 4 days after withdrawal of selegiline and totally abolished 7 days after discontinuation of the drug.& \textbf{Disease}: Orthostatic hypotension \textbf{Explanation}: a medical condition characterized by a sudden drop in blood pressure when standing up from a sitting or lying down position.
    \textbf{Confidence}: 0.9 & \textbf{Disease}: Orthostatic hypotension \textbf{Explanation}: Orthostatic hypotension is a type of low blood pressure that occurs upon standing up from a sitting or lying down position. It is mentioned in the sentence as a condition that was affected by the withdrawal and discontinuation of the drug.
    \textbf{Confidence}: 0.9  \\
    \Xhline{1.5pt}
    \end{tabular}}}
    \caption{Sample output predictions for sentences containing \colorbox{green!15}{target entities} ($T_E$)[\colorbox{red!15}{perturbed entities}($T'_E$)] in order to show the difference in the quality of explanations under zero-shot and few-shot setup. We only show predictions for the target entities. The model is equally confident in the prediction irrespective of the informativeness (more informative in few-shot) of the explanation. We show the explanations after perturbation in Table~\ref{tab:sample-after-pert}. }
    \label{tab:few-shot-explanations}
\end{table*}

\subsection{Manual Evaluation} \label{sec:manual} We manually evaluate explanations for a subset of examples in BC5CDR and CONLL.

\paragraph{Global vs. Local explanation.} Figure~\ref{fig:global-local-target} presents the change in the type of explanation under perturbation for target entities (see Figure~\ref{fig:global-local-other} for non-target entities). While, under zero-shot scenario, majority of the explanations are grounded in world knowledge (global) before and after the perturbation across all the perturbation types, we observe that $33\%$ of the explanations (BC5CDR) and $24.45\%$ (CONLL) change from global to local-level for Random perturbations showing that local context is required for predictions in such cases. E.g., the explanation generated for \textit{``Recently, we found that therapy with \textbf{r30s1k0} and L-dopa was associated with selective systolic orthostatic hypotension which was abolished by withdrawal of \textbf{r30s1k0}."} is ``\textit{R30s1k0 is a chemical compound used in therapy.}" while it was ``\textit{This is a medication used to treat Parkinson's disease}." before the perturbation. This connects to our zero-shot findings for automatic evaluation in ~\ref{subsec:auto}. Besides, we found that majority of the content in explanations generated under few-shot setup contains both globality and locality cues.

\paragraph{Prediction of target entities from context is easier for humans than ChatGPT. } While ChatGPT is able to correctly predict $67.8\%$ and $75\%$ of perturbed examples that are easy for humans for Alias, and Same Entity Type, respectively, only $45\%$, and $47\%$ are predictable for Random and Typo perturbations. This indicates that ChatGPT finds it harder to perform contextual predictions which are easier for humans. E.g., 
it is easy for humans to understand that the typo \textbf{legilinese} for \textbf{selegiline} is a chemical in \textit{``CONCLUSION: This study confirms our previous finding that \textbf{legilinese} in combination with L-dopa is associated with selective orthostatic hypotension."}, however, the model is not able to predict this entity. Besides, in case of random perturbation, it is easy for humans to understand from context that \textbf{8076mhq} is a type of disease in \textit{``METHODS: The cardiovascular responses to standing and head - up tilt were studied repeatedly in \textbf{8076mhq} patients receiving selegiline and as the drug was withdrawn."} however, ChatGPT fails to predict this entity. This shows that ChatGPT is not able to understand the type of entity from the context which humans can do easily.

\paragraph{Informativeness of the explanations change for the same non-target entity under perturbation.} We define the informativeness of explanations based on their source (world knowledge or from input) and amount of information, rated by a human on the basis of Likert Scale (1-5) where the input-grounded explanations are considered more informative than global explanations, and a combination of these as the most informative. E.g. when Israel is replaced with `Mount lebanon' in \textit{``Israel's Channel Two television said Damascus had sent a ``calming signal" to Israel.``}, the explanation for `Damascus' changes from `refers to the capital city of Syria' to less informative `refers to a geographical location'. 
%This is a medical condition characterized by a sudden drop in blood pressure when a person stands up from a sitting or lying down position." 
We observe (Figure~\ref{fig:informativeness}) that \% of least informative explanations (such as `France is a country') increases (or comparable) for person (disease) type in CONLL (BC5CDR) after perturbation for all types of perturbations while it decreases for location (drug) type except for \textbf{Typo} and \textbf{Random} in BC5CDR.
\section{Background and Related Work}
Pre-trained language models such as BERT \cite{devlin-etal-2019-bert}, BART \cite{lewis-etal-2020-bart}, etc., have shown their power to solve a wide variety of NLP tasks. Several large generative models have been proposed, such as GPT-3 \cite{NEURIPS2020_1457c0d6}, LaMDA \cite{thoppilan2022lamda}, MT-NLG \cite{megatron}, PaLM \cite{chowdhery2022palm}. LLMs usually exhibit amazing capabilities \cite{wei2022emergent} that enable them to achieve good performance in zero-shot and few-shot scenarios \cite{NEURIPS2022_8bb0d291, Wang2023ANT}.

Since ChatGPT does not reveal its training details, it imperative to evaluate privacy concerns; concerns that involve ethical risks \cite{haque2022i, krügel2023moral}, fake news \cite{jeblick2022chatgpt, chen-qian-2020-relation}, and financial challenges \cite{sun2023short, li2023chatgpt}. For its capabilities, researchers evaluate the performance of ChatGPT on different tasks, including machine translation \cite{peng2023making, jiao2023chatgpt}, sentiment analysis \cite{wang2023chatgpt}, question-answering \cite{guo2023close}, coreference resolution \cite{le2023large} and other NLP tasks \cite{bian2023chatgpt}. 
In addition, \cite{wei2023zeroshot} propose a two-stage framework, ChatIE, to use ChatGPT for zero-shot information extraction, and evaluate its performance on English and Chinese.

Several studies have been done to evaluate and improve the robustness of LLMs \cite{chen-durrett-2021-robust, awadalla-etal-2022-exploring, wang2021infobert,wang-etal-2022-measure}. Since this paper centers around the evaluation of robustness for NER tasks, it is worthy mentioning that prior researchers have assessed the NER model's robustness on token replacement \cite{bernier-colborne-langlais-2020-hardeval}, noisy or uncertain casing \cite{mayhew2019robust} and capitalization \cite{bodapati-etal-2019-robustness}. However, there has not been any comprehensive work in evaluating ChatGPT's robustness on NER and how the quality of explanations varies due to perturbations, which we are trying to fill up in this work.

\section{Conclusion}
We perform automatic and manual evaluation of explainability and IE capabilities of ChatGPT under the light of robustness before and after perturbations in the input. 
We find that ChatGPT is more brittle on domain-specific entity perturbations compared to the ones on widely known entities. 
Besides, we observe that the quality of explanations for the same entity \textit{considerably differ} under different types of perturbations and the quality can be significantly improved using in-context learning.
Even though ChatGPT is overconfident for incorrect predictions, its overconfidence can be significantly reduced using in-context learning.
To the best of our knowledge, we are the first to comprehensively analyze the effect of input perturbations on ChatGPT's predictions and the rationale behind its prediction on an IE Task.

% To the best of our knowledge, we are the first to comprehensively analyze the effect of adversarial perturbations on ChatGPT's predictions and rationale behind its prediction on an IE Task. We discover that ChatGPT is more brittle on rare entity perturbations compared to the ones on widely known entities.

\section*{Limitations}
While we analyze the faithfulness of the explanations with respect to the input, we do not evaluate if the global explanations are factual. During the manual evaluation, we observe that some of the perturbations resulted in invalid sentences or changed the meaning of the input, leaving this investigation for future work.

\section*{Ethics Statement}
Our method does not include any content that has potential risks or harms as we are analyzing the outputs of an existing model, ChatGPT. However, we acknowledge and condemn the malicious use of outputs of such AI systems to alter the opinions of the stakeholder and that these systems might generate biased outputs that need to be considered before using them for real-world applications.

\bibliography{anthology,custom}
\bibliographystyle{acl_natbib}

\appendix

\section{Example Appendix}
\label{sec:appendix}

\section{Prompt} \label{sec:prompt}
\subsection{Zero-Shot}
Identify named entities of type ``disease" or ``chemical" in the below text delimited by triple quotes. Format your response as a list of JSON objects with keys as ``type", ``entity", ``explanation", ``confidence" and values as ``type of the identified entity", ``identified entity", ``explanation of why it is an entity of that type",  and ``your confidence in identifying the entity as its type", respectively. Ensure that the identified entities can only be words or phrases present in the provided text. Confidence is a real value between 0 and 1.
    Text: ``````{text}"""

\subsection{Few-shot}
Your task is to identify the named entities of type "disease" or "chemical" in the given text delimited by triple quotes. Format your response as a list of JSON objects with keys as ``type", "entity", ``explanation", ``confidence" and values as ``type of the identified entity", ``identified entity", ``explanation of why it is an entity of that type", and "your confidence in identifying the entity as its type", respectively. Ensure that the identified entities can only be words or phrases present in the provided text. Confidence is a real value between 0 and 1. Use the following examples as a guide:
% Example 1:
% Example 2:
% Example 3:
% Example 4:
    
        EXAMPLE 1:
        Text: """None of the patients had decompensated liver disease""".
        Output: {{``entity": ``liver disease" , ``type": ``disease", ``explanation": ``It is a widely known disease and in the sentence it is mentioned that patients did not have decompensate this disease.",  ``confidence": 0.7 }}

        EXAMPLE 2:
        Text: """None of the patients had decompensated Measles.""".
        Output: {{``entity": ``Measles" , ``type": ``disease", ``explanation": ``Measles is a disease as it is a highly contagious, serious airborne disease caused by a virus that can lead to severe complications and death and in the sentence it is mentioned that patients did not have decompensate this disease.",  ``confidence": 0.9 }}

        EXAMPLE 3: 
        Text: """In conclusion , any disease can occur in patients receiving continuous infusion of 5 - FU."""
        Output: {{``entity": "5 - FU" , ``type": ``chemical", ``explanation": ``5 - FU is a chemical since it is a cytotoxic chemotherapy medication used to treat cancer and in the sentence it has been mentioned that any disease can occur because of its continuous infusion.",  ``confidence": 0.8 }}

        Example 4:
        Text: """In conclusion , any disease can occur in patients receiving continuous infusion of paracetamol. """    
        Output: {{``entity": ``paracetamol" , ``type": ``chemical", ``explanation": ``paracetamol is a chemical since it is a medication used to treat fever and mild to moderate pain and in the sentence it has been mentioned that any disease can occur because of its continuous infusion.",  ``confidence": 0.9 }}

        ======
        Text: """{text}"""
        Output:

\section{Additional Results} \label{sec:additional}
% \begin{figure}
%     \centering
%     \includegraphics[width=\columnwidth]{figs/correct_global_local_other.pdf}
%     \caption{Percentage of (input, perturbed input) pairs with change in type of explanations for non-target entities in BC5CDR.}
%     \label{fig:global-local-other}
% \end{figure}

\paragraph{Predicted entities may not be grounded in the input.} We observe a few predictions wherein the predicted entities are not even present in the input but are relevant given the context. E.g. ChatGPT predicts `schizophrenia' as one of the entities for \textit{``NRA0160 and clozapine antagonized locomotor hyperactivity induced by methamphetamine (\textbf{Hcxd8rf}) in mice."} as `clozapine' is used to treat schizophrenia.

\begin{table*}[!t]
\centering
    \tiny
    \resizebox{\textwidth}{!}{
    \scalebox{0.9}{\begin{tabular}{p{50mm} p{55mm} p{55mm}}
    \Xhline{1.5pt}
    \rowcolor{Gray}
     \textbf{Original Sentence[Perturbation]} & \textbf{Predictions After (Zero-shot)} & \textbf{Predictions After (Few-shot)} \\[1.1ex] \Xhline{1.5pt}
     % & \textbf{Predictions Before} & \textbf{Predictions After} \\
    \colorbox{green!15}{Selegiline}[\colorbox{red!15}{L-deprenyl}]-induced postural hypotension in Parkinson's disease: a longitudinal study on the effects of drug withdrawal.&  \textbf{Chemical}: L-deprenyl \textbf{Explanation}: L-deprenyl is a chemical substance used in the treatment of Parkinson's disease.
    \textbf{Confidence}: 0.9  & \textbf{Chemical}: L-deprenyl \textbf{Explanation}: L-deprenyl is a chemical since it is a medication used to treat Parkinson's disease and in the sentence it has been mentioned that it induced postural hypotension.
    \textbf{Confidence}: 0.9  \\
    \hline
    \colorbox{green!15}{Orthostatic hypotension}[\colorbox{red!15}{Orthostasis}] was ameliorated 4 days after withdrawal of selegiline and totally abolished 7 days after discontinuation of the drug.& \textbf{Disease}: Orthostasis \textbf{Explanation}: a medical condition characterized by lightheadedness or dizziness when standing up.
    \textbf{Confidence}: 0.8 & \textbf{Disease}: Orthostasis \textbf{Explanation}: Orthostasis is a medical condition where a person experiences lightheadedness or dizziness upon standing up. It is mentioned in the sentence that it was ameliorated after withdrawal of selegiline.
    \textbf{Confidence}: 0.8  \\
    \Xhline{1.5pt}
    \end{tabular}}}
    \caption{Sample output predictions for  sentences containing \colorbox{green!15}{target entities} ($T_E$)[\colorbox{red!15}{perturbed entities}($T'_E$)] in order to show the difference in the quality of explanations under zero-shot and few-shot setup. We only show predictions for the target entities. The model is equally confident in the prediction irrespective of the informativeness of the explanation.}
    \label{tab:sample-after-pert}
\end{table*}
% \paragraph{Explanation for the same predicted entity may be paraphrased before and after the attack.} 
% human eval 
% TODOS
% trigger-related metric.
% manual evaluation of incorrect predictions
% CONLL human eval
% aspect 
% -------------------------------------------------
% extra-predictions which are not gold-truth - nature of explanations
% Relation extraction perturbation survey
% Generate prompts for ICL for NER, prompt for aspect dataset
% Confidence calibration
% We need to get another API key for RC task
% Trigger matching algorithm 
% Generate results for NER for both zero and few-shot settings

\end{document}